\documentclass{article} 
\usepackage{iclr2017_workshop,times}
\usepackage{hyperref}
\usepackage{url}

\usepackage{amsmath}        
\usepackage{amssymb}        
\usepackage{flexisym}       
\usepackage{enumitem}
\usepackage{graphicx}
\usepackage{caption}
\usepackage{subcaption}
\usepackage{multirow}
\usepackage{float}

\DeclareMathOperator*{\argmax}{argmax}

\title{Revisiting stochastic off-policy action-value gradients}

\author{Yemi ~Okesanjo, Victor ~Kofia \thanks{} \\
Department of Computer Science\\
University of Toronto\\
Toronto, ON M5S 2E4, Canada \\
\texttt{\{o.okesanjo, victor.kofia\}@mail.utoronto.ca} \\
}

\begin{document}

\maketitle

\begin{abstract}
Off-policy stochastic actor-critic methods rely on approximating the
stochastic policy gradient in order to derive an optimal policy. One may also
derive the optimal policy by approximating the action-value gradient. The use
of action-value gradients is desirable as policy improvement occurs along the
direction of steepest ascent. This has been studied extensively within the
context of natural gradient actor-critic algorithms and more recently within
the context of deterministic policy gradients. In this paper we briefly
discuss the off-policy stochastic counterpart to deterministic action-value
gradients, as well as an incremental approach for following the policy
gradient in lieu of the natural gradient.
\end{abstract}

\section{Preliminaries}

\subsection{Stochastic off-policy theorem}

Consider a Markov decision process (MDP), where an agent receives a reward,
$r_{t}$, for an action, $a \in A$, taken in state, $s \in S$, according to
some stochastic behavioral policy, $b: S$ x $A \rightarrow (0, 1)$. We can
acquire a target policy, $\pi_{\theta}$, that maximizes the cumulative
rewards expected under this MDP by expressing its value as
\begin{align}
    J^{\pi}(\theta) = \mathbb{E}_{s \sim d^{b}}[V^{\pi}(s)] =
    \mathbb{E}_{s \sim d^{b}, a \sim b}[\varphi^{\pi, b}Q^{\pi}(s, a)]
\end{align}
where $V^{\pi}(s)$ is the expected cumulative rewards starting from a given
state, $s_{t}$, and $Q^{\pi}(s, a)$, the expected cumulative rewards,
starting from said state with an action, $a_{t}$; then following the policy,
$\pi_{\theta}$, until termination \citep{sutton:16}. The importance sampling
ratio, $\varphi^{\pi, b}$, scales $Q^{\pi}(s, a)$ according to the likelihood
of sampling the undertaken action from $\pi_{\theta}$, rather than $b$. In
order to find the parameters of $\pi_{\theta}$ such that the rewards are maximized, we can follow the direction of increasing performance
\begin{align}
    \nabla_{\theta} J^{\pi}(\theta) &= \mathbb{E}_{s \sim d^{b}, a \sim b}
    [\nabla_{\theta}\varphi^{\pi, b}Q^{\pi}(s, a) + \varphi^{\pi, b}
    \nabla_{\theta}Q^{\pi}(s, a)]\\
    &= \mathbb{E}_{s \sim d^{b}, a \sim b}\bigg[\varphi^{\pi, b}
    \frac{\nabla_{\theta}\pi_{\theta}(a|s)}{\pi_{\theta}(a|s)}Q^{\pi}(s, a)
    \bigg] + \mathbb{E}_{s \sim d^{b}, a \sim b}[\varphi^{\pi, b}
    \nabla_{\theta}Q^{\pi}(s, a)]\\
    &\approx \mathbb{E}_{s \sim d^{b}, a \sim b}\bigg[
    \varphi_{\pi, b}\frac{\nabla_{\theta}\pi_{\theta}(a|s)}
    {\pi_{\theta}(a|s)}Q^{\pi}(s, a)\bigg] = \widetilde{\nabla}_{\theta}
    J^{\pi}(\theta)
\end{align}

The first term in the above equation is the off-policy gradient and the second
term is the off-policy action-value gradient \citep{degris:12}. We want to
approximate the second term, so as to move in the policy gradient direction.

\subsection{Deterministic action-value gradient}

Action-value methods such as Q-learning acquire an implicit deterministic
policy, $\mu_{\theta}$, that can be expressed as
$a = \argmax_a Q^{\mu}(s, a)$. However this not feasible under continuous
action spaces. For such, the policy needs to be represented explicitly
\citep{silver:14, sutton:16}.

In order to learn the parameters for a continuous deterministic policy,
$\mu_{\theta}$, \citet{silver:14} proposed following the gradient of the
action-value, $\nabla_{\theta}Q^{\mu}(s, a)|_{a=\mu_{\theta}(s)}$, such that
the temporal difference (TD) error is minimized \citep{lillicrap:16}.
Using the chain-rule, if $Q^{\mu}(s, a)$ is a compatible function, the
gradient can be decomposed into the update equation
\begin{align}
    \theta_{t + 1} &= \theta_{t} + \alpha\mathbb{E}_{s \sim p^{\mu}}
    [\nabla_{\theta}\mu_{\theta}(s)\nabla_{a}Q^{\mu}(s, a)
    |_{a=\mu_{\theta}(s)} ]\\
    &= \theta_{t} + \alpha\mathbb{E}_{s \sim p^{\mu}}
    [\nabla_{\theta}\mu_{\theta}(s)\nabla_{\theta}\mu_{\theta}(s)^{\top}
    \omega]
\end{align}
where $\nabla_{\theta}\mu_{\theta}(s)$ is the deterministic policy gradient
and $\omega$ are the parameters of the action-value function that minimize
the TD error. The above equation moves in the same direction as the policy
gradient. We now discuss a similar case for stochastic policies.

\section{Stochastic Off-policy action-value gradient}
\label{actgrad}

\subsection{Compatible action-value functions}

In order to estimate how the parameters of an explicit policy change with
respect to $Q^{\pi}_{\omega}(s, a)$, the action-value needs to be compatible
with whatever type of policy is being represented. To do this, we
re-parametrize it as
\begin{align}
    Q^{\pi}_{\omega}(s, a) = A^{\pi}_{\omega}(s, a) + V^{\pi}_{\nu}(s)
\end{align}
where $A^{\pi}_{\omega}(s, a)$ is the advantage function of an action in a
state and $V^{\pi}_{\nu}(s)$ is the value of that state \citep{baird}. Due
to the zero-mean property of compatible features \citep{sutton:00,peters:05},
$Q^{\pi}_{\omega}(s, a)$ by itself cannot serve as a compatible function and
at the same time, a reliable estimator for cumulative expected rewards. In
practice, $A^{\pi}_{\omega}(s, a)$ is made compatible with respect to the
stochastic policy whilst $V^{\pi}_{\nu}(s)$ is used as a baseline. Following
\citep{sutton:00}, we represent the advantage function of a stochastic policy as
$A^{\pi}_{\omega}(s, a) =
\frac{\nabla_{\theta}\pi_{\theta}(a|s)}{\pi_{\theta}(a|s)}^{\top}\omega$.

\subsection{Stochastic off-policy gradient}

Given an action-value function, $Q^{\pi}_{\omega}(s, a)$, that is compatible
with the stochastic policy, $\pi_{\theta}$, in a manner shown above,
we can decompose the action-value gradient term in the complete off-policy
gradient theorem into
\begin{align}
    \mathbb{E}[\varphi^{\pi, b}
    \nabla_{\theta}Q_{\omega}^{\pi}(s, a)|d^{b}, b]
    &= \mathbb{E}_{s \sim d^{b}, a \sim b}[\varphi^{\pi, b}
    \nabla_{\theta}A_{\omega}^{\pi}(s, a)]\\
    &= \mathbb{E}_{s \sim d^{b}, a \sim b}[\varphi^{\pi, b}
    \nabla_{\theta}\pi_{\theta}(a|s)\nabla_{\pi_{\theta}}
    A_{\omega}^{\pi}(s, a)]\\
    &= -\mathbb{E}_{s \sim d^{b}, a \sim b}\bigg[\varphi^{\pi, b}
    \frac{\nabla_{\theta}\pi_{\theta}(a|s)}{\pi_{\theta}(a|s)}
    \frac{\nabla_{\theta}\pi_{\theta}(a|s)}{\pi_{\theta}(a|s)}^{\top}\omega
    \bigg]
\end{align}

where the squared log-likelihood gradients represent the Fishers information
matrix, $G^{\pi}(\theta)$. Consider the following from \cite{bhatnagar:09}

\textbf{Lemma 1}. The optimal parameters, $w^{\ast}$, for the compatible
function of a stochastic policy, $\pi_{\theta}$, can be expressed as
\begin{align}
    \omega^{\ast} =
    G^{\pi}(\theta)^{-1}\widetilde{\nabla}_{\theta}J^{\pi}(\theta)
\end{align}

which represents the natural policy gradient \citep{amari:98, kakade:02}. Hence
the above moves in the off-policy gradient direction as both squared terms in
the action-value gradient cancel out.

\section{Experimental results}
\label{experiments}

We evaluate the performance of an actor-critic algorithm that follows the
policy gradient based on the above equation. We compare this algorithm,
Actgrad, alongside two other algorithms: the off-policy actor-critic
(Offpac) and Q-learning (Qlambda). Experiments are performed on the Cart
Pole and Lunar Lander environments provided in the Open AI gym
\citep{brockman:16}.

\subsection{Details}

The first task we consider, Cart Pole, is the task of balancing a pole
attached atop a cart by an un-actuated joint. The goal is to apply a force,
F, to either the right or left of the pole in order to keep it upright. For
each time step the pole is upright, a reward of 1 is given.

Next we consider, Lunar Lander, the task of piloting a lander module through
the lunar atmosphere and unto a landing pad at the center of the screen. The
goal is to bring the spacecraft to rest by either doing nothing or firing the
main, left or right engines. Bad landings incur negative rewards, as do
firing the engines. However, larger rewards are given for smooth landings.

For Cart Pole, the state features are encoded using the Boxes method
\citep{barto:83}, while for Lunar Lander they are encoded using heuristics
provided by Open AI. Agents were trained on the environments for 1500 \& 700
episodes respectively, with the same training parameters \& learning rates
shared across them. On Cart Pole, training episodes ended after 250 time
steps while on Lunar Lander, they ended after 500 time steps. Training was
repeated for 10 trials on each environment and testing was performed for
100 episodes after each trial.

\subsection{Results}

We now present the training and test results for the evaluated algorithms
on each environment.

\begin{table}[H]
\caption{Average Test Results.}
\label{sample-table}
\begin{center}
  \begin{tabular}{|l|l|l|l|l|}
    \hline
    \multirow{2}{*}{Agent} & \multicolumn{2}{c|}{Cart Pole} &
    \multicolumn{2}{c|}{Lunar Lander}\\
    & Rewards & Episodes Solved & Rewards & Episodes Solved\\
    \hline
    Offpac   & 214.56 $\pm$ 3.78 & 99.9\% & 152.89 $\pm$ 8.52 & 91.7\%\\
    \hline
    Qlearning & 167.12 $\pm$ 2.97 & 99.0\% & 109.30 $\pm$ 8.87 & 79.2\%\\
    \hline
    Actgrad  & 209.18 $\pm$ 3.68 & 99.7\% & 109.46 $\pm$ 11.27 & 85.7\%\\
    \hline
  \end{tabular}
\end{center}
\end{table}

\begin{figure}[h]
\begin{subfigure}{.50\textwidth}
\begin{center}
\includegraphics[width=1.0\textwidth]{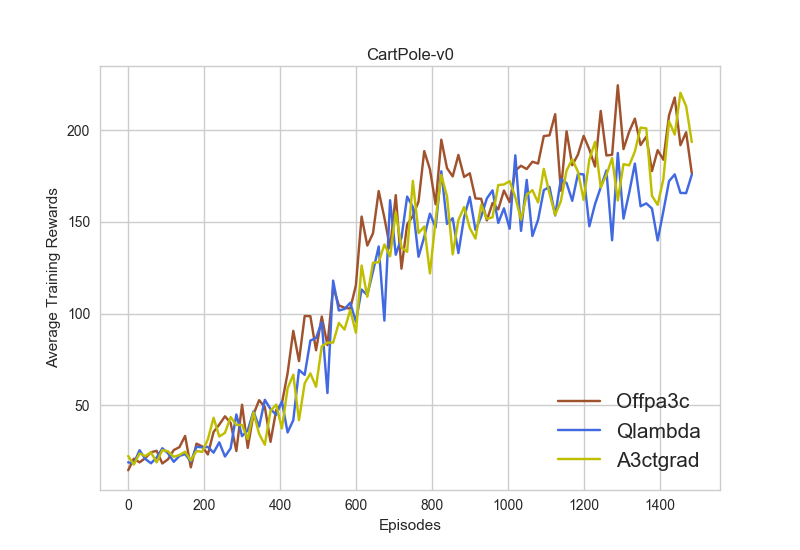}
\end{center}
\caption{Cart Pole.}
\end{subfigure}
\begin{subfigure}{.49\textwidth}
\begin{center}
\includegraphics[width=1.0\textwidth]{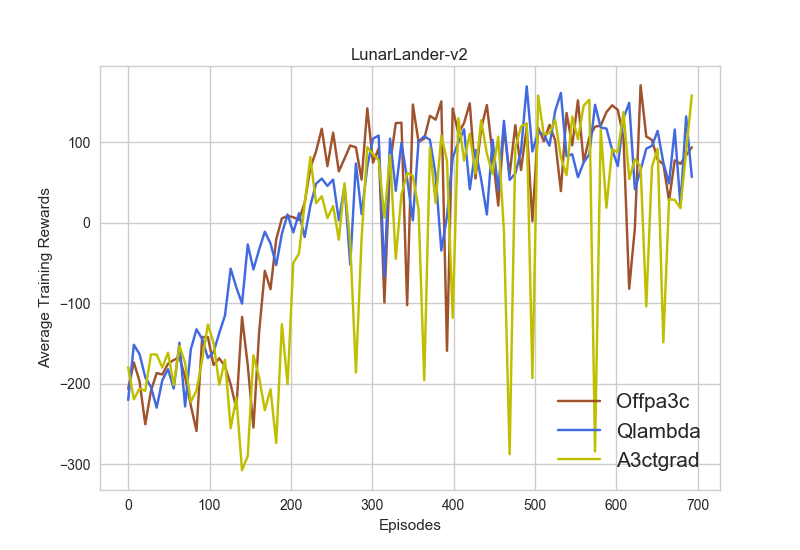}
\end{center}
\caption{Lunar Lander.}
\end{subfigure}
\caption{Average Training Rewards.}
\end{figure}

From the above, the investigated algorithm reaches a training performance
close to that of the off-policy actor-critic on Cart Pole. However it suffers
from higher variance on the Lunar Lander environment. This may be due to the
fact that learning relies on estimates of the advantage as determined by the
current advantage parameters, rather than the actual advantage gotten from
the critic. This is likely pronounced due to the difficulty of Lunar Lander in comparison to Cart Pole.

\section{Discussion}
\label{discussion}

In this paper, we have discussed a method for following the stochastic
off-policy gradient in a manner similar to that of the deterministic policy
gradient. We then compared the performance of such method with other policy
gradient algorithms. Although the approach suffers from high variance on
certain tasks, it nevertheless outperforms deterministic algorithms and can
easily be made to follow the steepest ascent direction by dropping the
natural gradient term.

\bibliography{iclr2017_workshop}
\bibliographystyle{iclr2017_workshop}
\nocite{*}

\end{document}